\documentclass[10pt,twocolumn,letterpaper]{article}
\usepackage[accsupp]{axessibility}
\usepackage{cvpr}
\usepackage{times}
\usepackage{epsfig}
\usepackage{graphicx}
\usepackage{svg}
\usepackage{amsmath}
\usepackage{amssymb}
\usepackage{multirow}
\usepackage{array}
\usepackage{tabularx}
\usepackage{caption}
\usepackage{subcaption}
\usepackage{hyperref}
\usepackage{gensymb}
\hypersetup{
    colorlinks=true,
    linkcolor=blue,
    filecolor=magenta,      
    urlcolor=cyan,
    pdftitle={Overleaf Example},
    pdfpagemode=FullScreen,
    }

\newcolumntype{P}[1]{>{\centering\arraybackslash}p{#1}}

\begin{document}

\title{Recent Advancement in 3D Biometrics using Monocular Camera}


\author{
Aritra Mukherjee, Abhijit Das\\
Birla Institute of Technology and Science, Pilani – Hyderabad Campus \\
Secunderabad, Telangana 500078\\
{\tt\small abhijit.das@hyderabad.bits-pilani.ac.in}
}

\maketitle
\thispagestyle{empty}

\begin{abstract}
Recent literature has witnessed significant interest towards 3D biometrics employing monocular vision for robust authentication methods. Motivated by this, in this work we seek to provide insight on recent development in the area of 3D biometrics employing monocular vision. We present the similarity and dissimilarity of 3D monocular biometrics and classical biometrics, listing the strengths and challenges. Further,
we provide an overview of recent techniques in 3D biometrics with monocular vision, as well as application systems adopted by the industry. Finally, we discuss open research problems in this area of research.

\end{abstract}

\section{Introduction}
\begin{figure}[t]
\begin{center}

   \includegraphics[width=\linewidth]{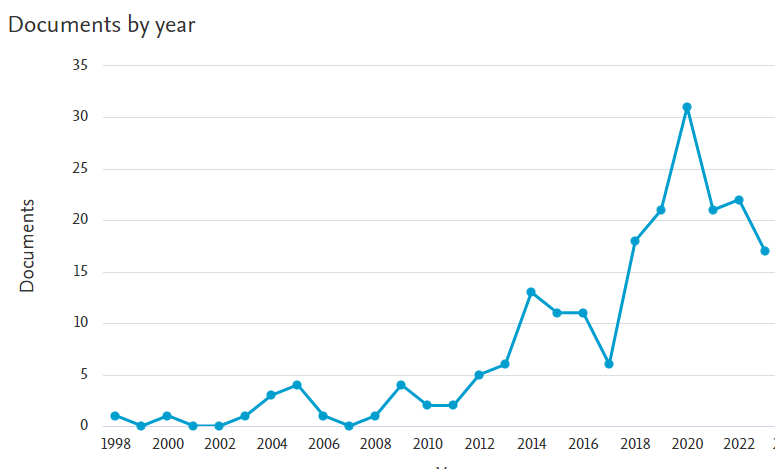}
\end{center}
\vspace{-2mm}
   \caption{Graph of scientific papers published per year in the period 1998-2022 matching keywords including “3D biometrics” and "single camera monocular vision”. From Scopus (https://www.scopus.com).}
\label{pub}
\vspace{-4mm}
\end{figure}

In the present times, biometrics sensors and systems have become a part of daily human life across multiple spheres, where some kind of security is needed. From applications like digital financial transactions to border security, the role of various biometrics is unquestionable. Despite of all such advancements many forms of biometrics still needs research attention w.r.t to robustness and preciseness. Further, most of the existing biometric systems like fingerprints and palm prints, are touch-based systems. With the episode of COVID-19 the world has learnt how such systems can be rendered useless. To mitigate the same many expensive sensor technologies such as 3D sensors and other contact-less biometrics has been proposed. Most of them are expensive, hence affordability is a big question for the development of low-cost systems. In these contexts, the use of monocular camera as the alternative hardware, have gained steam in recent times (See Figure~\ref{pub}). Also, several commercial solutions have been released\cite{1}. Among all the biometrics, face leads the race with more than half of the existing technologies related to 3D biometrics based on monocular vision, followed by gait, finger, ear, eye movement and vein. In many cases, a monocular camera is supplemented with other specialized low-cost emitters like IR LEDs, or are motorized. In almost all these works, some form of 3D information retrieval is involved. Hence, in this review, we have tried to give an overview of such emerging technologies.

As mentioned, most of the advances are in the direction of 3D biometrics with face \cite{samatas2022biometrics} followed by fingerprint \cite{rusia2023comprehensive}, presentation attack detection \cite{hernandez2023introduction}. Along with face biometrics ear \cite{ganapathi2023survey} is used for multimodality. 
3D vascular biometric\cite{iula20223d} and palm based biometric\cite{wang2023anti} has recently gained steam in this avenue of research. Thus in many forms of human biometrics, 3D information is crucial. But the need for sophisticated and expensive hardware components works as a deterrent to its widespread adoption, as a result monocular vision for 3D data came into existence. 

Most of the modalities used in 3D monocular biometrics employ Structure from shadow (SFS) and Structure from motion (SFM) to harness 3D information \cite{samatas2022biometrics}. There are approaches based on monocular vision\cite{li20223d3m,zhu2022beyond,tewari2022disentangled3d} but still those methods fail under drastic lighting conditions. Thus recent advancement\cite{tiwari2022occlusion} has worked in that direction. In recent times monocular 3D object reconstruction has made a lot of improvements due to the neural radiance field approach or NeRF\cite{gao2022reconstructing,khan2022towards,sun2022fenerf,deng2023learning} and the area needs more research for efficiency and robustness in applications.  Observing the recent trends, monocular image-based 3D biometrics is an emerging topic that has myriad applications and the best part is, it is much more affordable. Hence, it will be beneficial to document the recent development in this area. 

Therefore, in the following sections, we are going to discuss the benefits and challenges of 3D biometrics using monocular vision in comparison to traditional biometrics. Followed by a critical review of recent advancements in this subject, categorized into different biometric traits, namely, face, gait, vein pattern, eye movement and others. Though the majority of the works reviewed are based on monocular images or cameras, some of them have used other expensive sensors. Next, we will discuss the way forward. 

\section{3D biometrics with monocular vision}
A classical biometric system acquires data from an individual (e.g., a face image), processes and mathematically models it, and compares mathematical representation with the distribution of a population in order to verify a claimed identity or to determine an identity. Keeping in mind that this classical structure is also incorporated in the context of 3D biometrics with monocular vision systems, one could introduce the following definition. \textit{A 3D biometrics with a monocular vision system is a special case of 3D biometrics that uses monocular sensors to get leverage that a traditional 3D biometrics achieves with sophisticated sensors.} Hence, such a system can be low cost reliable biometric technology. Now we proceed to enlist the benefits and challenges in 3D biometrics with monocular vision.

\subsection{Benefits}
\begin{figure}
    \centering
    \includegraphics[width=68mm]{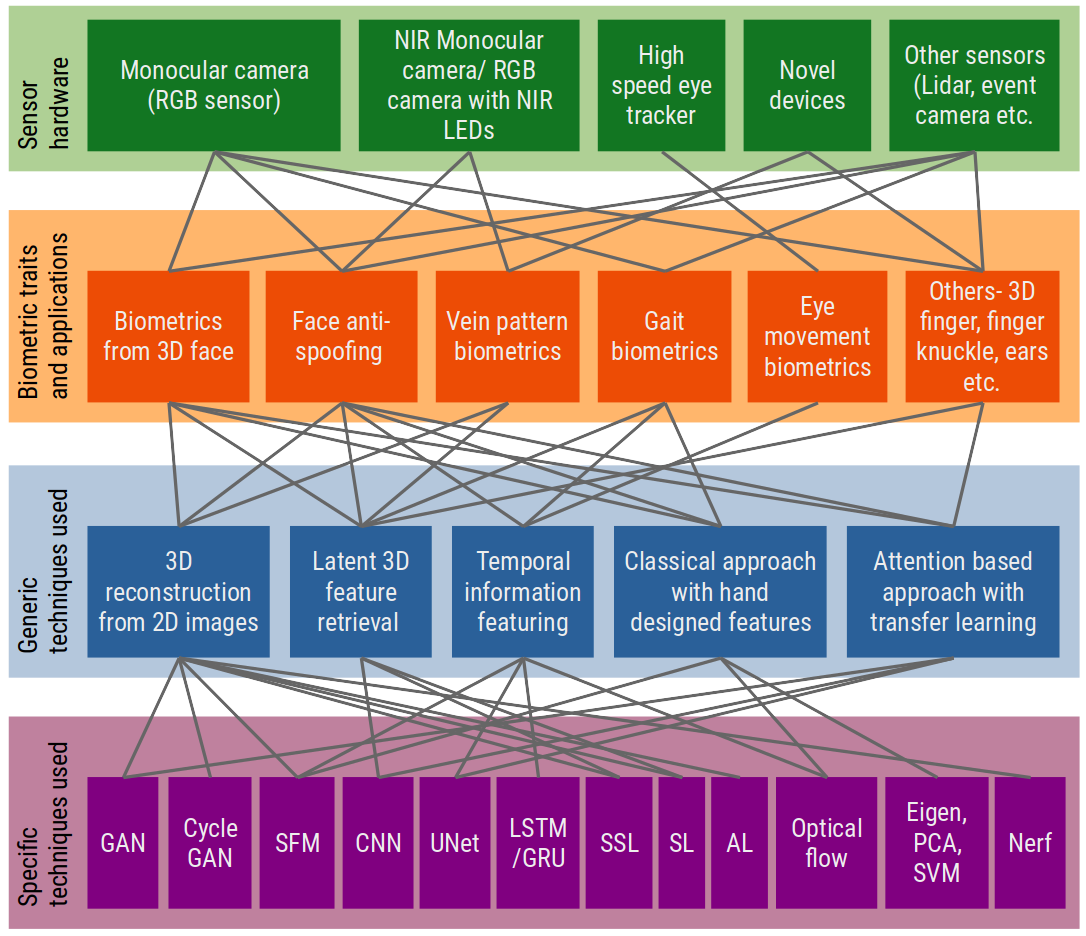}
    \caption{Relationship between sensor hardware, biometric traits and applications and methodologies used for them, as per recent development in the field. }
    \label{fig:graph_survey}
\end{figure}

The future of biometrics technology is slowly turning towards the use of deep learning models using images or videos obtained from normal monocular cameras as studied in this work~\cite{imaoka2021future}. Figure~\ref{fig:graph_survey} shows the relationship among different hardware sensors, biometrics traits and methodologies in recent works. It is to be noted that the abundance of 3D techniques and the use of the monocular camera is quite significant. There are many obvious reasons for this which can be broadly classified into four points,i.e.
\begin{itemize}
    \item \textbf{Ease of sensor usage:}  Fast growth of hardware technology has made deep learning available on edge devices with minimal capital and operating expenditure. The abundance in the availability of monocular cameras across price ranges and easy amalgamation of existing infrastructure also helps in ease of use. 
    \item \textbf{Low cost:} Monocular camera still remains as one of the cheapest sensor technologies with a very high information bandwidth. 
    \item \textbf{Covertness: }The other advantage is contact-less usage and sometimes, as required by scenarios, covert acquiring of data. 
    \item \textbf{Operational simplicity:} Other than cost-related benefits, the simplicity of servicing biometric devices with a monocular camera cannot be argued, which is making it a primary choice for sensors. 
\end{itemize}

\subsection{Challenges}
There are apparently some challenges attached to the usage of monocular cameras as the only sensor hardware. The works discussed in this review sheds light on them when used for various traits, especially in Section~\ref{lab_vein} and~\ref{lab_eye}. The major challenges can be summarized as follows:
\begin{itemize}
    \item \textbf{Proper spectrum band: }Many biometric traits rely on information that is unobtainable until a monocular camera is supplemented with special emitters or filters. Especially for subcutaneous information like vasculature and vein pattern IR LEDs are needed and sometime NIR filters are needed. That takes the overall cost of the system higher.
    \item \textbf{Unavailability of direct 3D data: } Some sort of algorithms, either classical or machine learning oriented, are needed to make a pseudo depth map properly out of the data from the monocular camera. 
    \item \textbf{Dependence on lighting model: }Many works on biometrics using monocular camera needs structured or directed lighting for proper featuring of visual information, especially 3D features. This makes the usage of monocular cameras restricted to certain environments.
    \item \textbf{Calibration of camera parameters: }Many parameters of a camera such as auto-focus, white balance, ISO, exposure, shutter speed and color model are auto-calibrated for using monocular cameras in other common commercial usages. Only industrial-grade expensive cameras have the flexibility of tuning these parameters by the algorithm in feedback mode for efficient information sampling. This poses a challenge for making deep learning models more robust to handle these fluctuations. 
\end{itemize}

\section{Recent advancement on 3D biometrics
with monocular vision}
In this section, we do a critical review of the most recent and innovative work on 3D biometrics with monocular vision. Apart from biometric recognition itself, several other aspects such as presentation attack detection, and virtual reality (VR) that have been investigated by the research community are also discussed. Additionally, this section also highlights how the identified challenges have been addressed in the recent literature and the commercial solutions available.

\subsection{Face biometrics}
Among all biometrics traits, the face is the most popular one \cite{samatas2022biometrics}. There had been many surveys on face recognition methodologies through the last decade~\cite{jafri2009survey,serrano2010recent,galbally2014biometric,wang2014comprehensive,rattani2018survey} which show the use of traditional techniques like hand-designed feature, eigenfaces, Gabor filters and visual word for most of the approaches. From the beginning of the present decade, surveys recognized the booming of deep learning techniques in face biometrics~\cite{taskiran2020face,minaee2023biometrics} and the need for privacy in face biometrics~\cite{meden2021privacy}. Recent works in this direction have been briefly described in Table~\ref{tab_3d_face}. The Table~\ref{tab_face_antispoof} discusses about recent developments in face anti-spoofing techniques using monocular images. Some works recognized the need for 3D face recognition along with deep learning~\cite{kaur2020facial,jia2020survey,li2022comprehensive,sharma20223d} due to various emerging challenges. When it comes to reliable face recognition for biometrics a number of challenges crops up in real-world applications such as demographic bias~\cite{drozdowski2020demographic}, cancellable biometrics~\cite{manisha2020cancelable}, face beautification~\cite{rathgeb2019impact,rathgeb2020makeup}, face masking~\cite{gomez2022biometrics} and face morphing~\cite{venkatesh2021face}.


Taking all such constraints and challenges into consideration, we have focused our review on 3D face-based biometrics as it can dilute many of the above-highlighted challenges. Most of these works use a monocular camera. These works attempt to create a pseudo depth map~\cite{khan2023robust,malah20233d} or a 3D morphable model~\cite{han2022accurate,li20233d} of the face from a single image or multiple images captured with a static camera and a moving head. GAN-based approaches~\cite{jin2022pseudo,zeng2022joint,malah20233d,khan2023robust,chen2023recognition} are used in most of the approaches for extracting 3D information. An early work~\cite{han2020can} argued in favour of using 3D biometrics from 2D face images as it is more robust to pose and lighting variances. Applications like Deepfake detection are also been explored in this direction. Depthfake~\cite{maiano2022depthfake}, which concludes that deep fake detection is more robust when 3D reconstruction is done from 2D. For the face, anti-spoofing detection of temporal repetitive patterns~\cite{verissimo2023transfer,yu2020searching,muhammad2022self} and classification of materials by its irradiation property~\cite{li2022image,hassani2023monocular}, are the two most common trend used on short monocular facial video clips.  

Further, the importance of 3D face decomposition from 2D images for biometrics, especially forgery detection has been well discussed in a recent work~\cite{zhu2023face}. It elaborates how by decomposing a 2D image into graphics components including 3D shape, lighting, common texture, and identity texture the task of biometric recognition becomes a lot more robust.
The use of NeRF for constructing 3D face avatar\cite{xu2023latentavatar} is another approach to creating photorealistic 3D faces from 2D portraits with cross personality traits i.e. facial features and structure from different individuals. 

\begin{table*}[]

\caption{Recent works in 3D face reconstruction and biometrics monocular camera}
\label{tab_3d_face}
\resizebox{\textwidth}{!}{%
\begin{tabular}{|p{20mm}|p{20mm}|p{50mm}|p{45mm}|p{30mm}|p{20mm}|}
\hline
\textbf{Work} & \textbf{Sensor} & \textbf{Technique} & \textbf{Dataset} & \textbf{Accuracy} & \textbf{Challenge} \\ \hline
Jin \etal~\cite{jin2022pseudo} &
  normal monocular camera &
  shearlet transform combined with a generative adversarial network (GAN), multi conditional image to image translation &
  Bosphorus 3D Face Database~\cite{savran2008bosphorus}, CASIA 3D~\cite{CBSR} for training, BU-3DFE~\cite{yin20063d} for testing &
  SSIM $0.869$, RMSE $23.99$, PSNR $20.53$ &
  Occlusion \\ \hline
Deng \etal~\cite{deng2022fast} & normal monocular camera &  Attention mechanism with graph convolutional network, rough 3D data to train the model & LFPW, HELEN, IBUG and XM2VTS combined by Guo~\etal~\cite{guo2020towards} with face landmarks generated from FAN~\cite{bulat2017far} for training and NoW dataset~\cite{sanyal2019learning} & reconstruction error median $1.29$mm, mean $1.63$mm std $1.41$mm & Albedo, speed vs accuracy for smaller parameter size\\ \hline
 Zeng\etal~\cite{zeng2022joint} & normal monocular camera & Joint reconstruction by a deep shape reconstruction and texture completion network, U-V texture map and inpainting network, explainable &  Florence hybrid face dataset~\cite{bagdanov2011florence} & EER $6.39\%$, mean RMSE $2.09$mm & occlusion, non-face objects non-symmetric face \\ \hline
 Han \etal~\cite{han2022accurate}& normal monocular camera & roughly generated 3D morphable model (3DMM) using a pixel level microfacet estimation & No dataset other than trained 3DMM & The RMSE $2.707$mm & works on non uniformly lit faces only\\ \hline Angermann~\etal~\cite{angermann2022unsupervised} & normal monocular camera & Cyclic optimization of RGB-to-depth and depth-to-RGB networks for unsupervised single-shot depth estimation & Texas 3DFRD~\cite{gupta2010texas}, SURREAL~\cite{varol2017learning} and Bosphorus 3D Face Database~\cite{savran2008bosphorus} &  RMSE $\pm$ std $0.068 \pm 0.0.27$, MAE $\pm$ std $0.051 \pm 0.023$ & four neural networks must be fitted in parallel \\ \hline
 Kang~\etal~\cite{kang2022facial}& Dual pixel monocular camera & Network with two novel modules, namely, Adaptive Sampling Module and Adaptive Normal Module, generalized depth and normal estimation & Public DP dataset~\cite{abuolaim2020defocus} and captured dataset $135744$ face images for $101$ subjects with canon DSLR (focus distance $1.0$ to $1.5$ m) & WMAE $0.085$ WRMSE $0.133$ & Fixed focus distance, bias to skin tone and lighting \\ \hline Malah~\etal~\cite{malah20233d}& normal monocular camera & GAN and Graph CNN based approach with facial landmark location in the 2D face image as input and 3D mesh as output & AFLW2000-3D dataset~\cite{zhu2017face} and Florence hybrid face dataset~\cite{bagdanov2011florence} & AFLW2000-3D EMD $0.13$ Chamfer $0.0077$, Florence EMD $0.269$ Chamfer $0.0623$ & Biased to texture and lighting, not good around eyes \\ \hline Li~\etal~\cite{li20233d}& normal monocular camera  & 3DMM refinement by end to end training by residual learning, combination of low level losses with fine-grained optimization, high-quality 3D faces from moderate-quality 2D faces & 
 CelebA~\cite{liu2015deep}, 300WLP~\cite{zhu2016face}, LS3DW~\cite{bulat2017far}, LFW~\cite{huang2008labeled}, FFHQ~\cite{karras2019style} and IJB-C~\cite{maze2018iarpa} for training Florence~\cite{bagdanov2011florence} NoW~\cite{sanyal2019learning} Facewarehouse~\cite{cao2013facewarehouse} for testing & 3D-RMSE (in mm) on NoW mean $1.9$ median $1.52$ std $1.49$ Florence mean $2$ median $1.55$ std $1.57$ Facewarehouse mean $1.91$ & not good for Asian faces \\ \hline Kao~\etal~\cite{kao2023towards} & normal monocular camera & Perspective network, a novel deep network architecture learns 3D to 2D face coordinates and then uses the reverse to estimate very accurate 3D faces from 2D images & ARkit (contributing dataset) $400$ people $717840$ samples for training $100$ $184884$ samples for testing, BIWI~\cite{fanelli2013random} & RMSE (mm) ARkit mean $1.76$ median $1.72$ & fails under high occlusion \\ \hline Khan~\etal~\cite{khan2023robust} & normal monocular camera and synthetic monocular camera & Lightweight feature fusion model that used an encoder-decoder type architecture pretrained on synthetic data, small mode huge data & Pandora~\cite{borghi2017poseidon}, Eurecom Kinect~\cite{min2014kinectfacedb}, BIWI~\cite{fanelli2013random}, Synthetic human facial depth~\cite{khan2021efficient} & Synthetic human facial depth RMSE $0.023$ & training requires huge dataset for optimal results\\ \hline Chen~\etal~\cite{chen2023recognition} & normal monocular camera & Well-trained CycleGAN with 3DMM for pretraining with dentical cycle consistency loss and two perceptual losses with identity preservation in 2D and 3D & ND-2006~\cite{faltemier2007using}, CASIA 3D~\cite{CBSR}, 3D-TEC~\cite{vijayan2011twins}, Bosphorus 3D Face Database~\cite{savran2008bosphorus}, Texas 3DFRD~\cite{gupta2010texas} & Rank1 recognition accuracy 3D-TEC $0.5$, ND-2006 $0.872$, CASIA 3D $0.706$, Bosphorus $0.641$, Texas-3FRD $0.925$ & high pose variation\\ \hline
 Mohaghegh~\etal~\cite{mohaghegh2023reinforced} & normal monocular camera & Reinforced Active Learning for dynamically selecting the most informative view-points by clustering based pooling  & 300WLP, AFLW-2000~\cite{zhu2016face} & for similar frameworks, it needs $40\%$ of the labelled data & informative view-points mandatory in training\\ \hline
 
\end{tabular}%
}
\end{table*}

\begin{table*}[]

\caption{Recent works in face anti-spoofing using monocular camera}
\label{tab_face_antispoof}
\resizebox{\textwidth}{!}{%
\begin{tabular}{|p{20mm}|p{60mm}|p{30mm}|p{50mm}|p{20mm}|}
\hline
\textbf{Work}  & \textbf{Technique} & \textbf{Dataset} & \textbf{Accuracy} & \textbf{Challenge} \\ \hline
 Lin~\etal~\cite{lin2022distributed} & Face anti-spoofing based distributed learning, learns semantic relationship by Eigen decomposition of manifold matrix with hypergraph Laplacian manifold learning in a distributed way & MSSPOOF~\cite{chingovska2016face}, CASIA-SURF~\cite{zhang2012face} & FN on MSSPOOF $0.14$, FP on MSSPOOF $0.018$, FN on CASIA-SURF $0.17$, FP on CASIA-SURF $0.012$ & High training data amount \\ \hline
 Verissimo~\etal~\cite{verissimo2023transfer} & Transfer learning on pretrained VGG16 & Replay-Attack dataset~\cite{chingovska2012effectiveness}, OULU-NPU~\cite{boulkenafet2017oulu}, NUAA~\cite{tan2010face}, MSU-MFSD~\cite{wen2015face} & ERR(in $\%$ Raw an Subsampled) NUAA $0, 0$, Replay-attack $0.67, 1.98$, MSU-FASD $5.32, 5.36$, OULU $4.69, 7.18$ & May fail on 3d face mask PA\\ \hline
 Singh~\etal~\cite{singh2022fusion} & Pair-wise pose normalization on the real and fake face with global affine alignment, signed difference between AlexNet and ResNet enrolled and checked with SVM for verification & Morph ABC dataset~\cite{singh2019robust} & D-EER $6.18 \pm 3.7$, BPCER20 $8.3 \pm 0.1$, BPCER10 $4.2 \pm 0.1$ & Does not handle non-rigid deformation\\ \hline
 Yu~\etal~\cite{yu2020searching} & Central Difference Convolutional Network with Neural Architecture Search with Multiscale Attention Fusion Module, aggregation of intensity and gradient information for capturing intrinsic detailed pattern & OULU-NPU~\cite{boulkenafet2017oulu}, MSU-MFSD~\cite{wen2015face}, Replay-Attack dataset~\cite{chingovska2012effectiveness}, SiW~\cite{liu2018learning}, SiW-M~\cite{liu2019deep}, CASIA-SURF~\cite{zhang2012face} & Overall AUC $96.63  
 \pm 9.15\%$, Train on CASIA, test on Replay HTER $6.5\%$, Train on Replay test on CASIA HTER $29.8\%$ & not context aware, not cross application \\ \hline
 Muhammad~\etal~\cite{muhammad2022self} & novel video preprocessing named Temporal Sequence Sampling, SSL on CNN by removing interframe affine transforms for liveliness detection & Replay-Attack dataset~\cite{chingovska2012effectiveness}, OULU-NPU~\cite{boulkenafet2017oulu}, MSU-MFSD~\cite{wen2015face}, CASIA-SURF~\cite{zhang2012face} & Train on CASIA, test on Replay HTER $5.9\%$, Train on Replay test on CASIA HTER $15.2\%$, train on MSU-MFSD and cross dataset HTER (Replay, CASIA) $28.66\%$ , OULU-NPU included $30.12\%$ & needs at least 2 seconds clip \\ \hline
 Li~\etal~\cite{li2022image} & Attentional face image to 1D CNN for extracting points with the absorption of light based on blood vessel underneath, then GRU to learn temporal variation in absorption & MSU-MFSD~\cite{wen2015face}, Replay-Attack dataset~\cite{chingovska2012effectiveness}, 3DMAD~\cite{erdogmus2014spoofing} & 3DMAD EER $0.29\%$, MSU EER-P $0.97$, EER-R $16.55$, Replay attack EER-P $2.18\%$ HTER-P $1.60\%$ EER-R $8.52$, HTER-R $10.03\%$ & tone biased, needs good lighting \\ \hline
  Hassani~\etal~\cite{hassani2023monocular} & Near-infrared spectroscopy with visible light image correspondence for PAD, using MobileV3 net & collected dataset of $30$ individuals with $80000$ unique frames & ACER, NPCER $0.2$, $0.2$ in lab condition, $0.3$, $0.4$ in exterior lighting & Needs IR camera with RGB \\ \hline

\end{tabular}%
}
\end{table*}

\subsection{Vein pattern biometrics}
\label{lab_vein}
The fact that subcutaneous vein pattern can be used as a biometric trait, was established by earlier works~\cite{adeoye2010survey,jain2010biometrics},  which also prophesied the popularity of the trait. Recently vein pattern biometrics has received significant attention and multiple approaches for that have been explored well in the literature. They include both hardware and algorithm development, which are covered in some recent surveys~\cite{wu2020review,uhl2020state,jia2021survey,al2022vein}. After going through existing surveys we discovered that there is no consolidated work that reviews recent developments in the field using monocular RGB and IR cameras. In Table~\ref{tab:tab_vein_pattern} some of the recent works in 3D vein pattern using monocular camera has been briefly discussed.  Some of the works are focused on specialized hardware development~\cite{qin2022local,zhao2022exploiting,kauba2022three} for quick scanning of palm and finger veins, while some have taken periocular vasculature~\cite{bhattacharya2022portable} as the source of subcutaneous vein pattern. CNN is used as the common methodology in most approaches~\cite{zhao2022exploiting,bhattacharya2022portable}.  Most of the work has contributed by developing challenging datasets. 

\begin{table*}[]

\caption{Recent works in vein pattern as biometric using a monocular camera in RGB and NIR modes}
\resizebox{\textwidth}{!}{%
\label{tab:tab_vein_pattern}
\begin{tabular}{|p{18mm}|p{20mm}|p{60mm}|p{50mm}|p{20mm}|p{15mm}|}
\hline
\textbf{Work} & \textbf{Sensor} & \textbf{Technique} & \textbf{Dataset} & \textbf{Accuracy} & \textbf{Challenge} \\ \hline
 Qin~\etal~\cite{qin2022local} & Motorized monocular camera and IR LEDs & Novel device, Rotated camera captures multiple images subjected to Unet for ROI extraction and local attention transformer stitches local vein patterns for classification & Dataset collected from $80$ subjects with $1280$ sequences ($8 \times$ sequences $2 \times$ fingers $80$ subjects) with $76800$ frames ($60$ frames $\times 1280$ sequences) & $99.80\%$ on single view and $91.25\%$ to $97.19\%$ for multiview  & requires specialized hardware \\ \hline
 Zhao~\etal~\cite{zhao2022exploiting}& Monocular camera with IR LEDs & Novel device, Hierarchical Content-Aware Network composed of Global Stem Network (lightweight CNN + bidirectional gated GRU) and Local Perception Module with novel entropy loss extracts discriminative hierarchical features for classification & Multi-perspective Finger Vein dataset (THU-MFV) collected $5400$ images from $40$,  participants divided in A1 $100$ classes, A2 $50$ classes with over/underexposed images, Large Multi-View Finger Vein database (LMVFV) $660$ classes from $176$ participants with $23760$ images & THU-MFV EER A1 $0.97\%$, A2 $1.55\%$, LMVFV EER $1.22\%$ & requires specialized hardware \\ \hline
 Kauba~\etal~\cite{kauba2022three} & Motorized NIR camera with mirrors and NIR LEDs & Novel device with NIR sensitive camera Basler acA4112-30uc with lens $25$ mm/F$1.8$ from Edmund Optics, $805$ nm to $811$ nm laser diodes, $75\degree$ Powell lens, $4$ mirrors & Extension of SCUT-3DFV-V1~\cite{kang2019study} to $702$ distinct fingers & Not applicable as work is on a novel hardware design only & images not sharp, not all angles covered\\  \hline Bhattacharya ~\etal~\cite{bhattacharya2022portable} & NIR camera with Raspberry Pi 4 Model B & Vein and Periocular Pattern based Convolutional Neural Network that fuses forehead Vein pattern and Periocular pattern embedding for classification, Gamma correction, Difference of Gaussian and Contrast enhancement used as preprocessing & contributed Forehead Subcutaneous Vein Pattern (FSVP) and Periocular Biometric Pattern (PBP) dataset FSVP-PBP with $200$ subjects consisting $2000$ images & EER $0.08$ accuracy $98.6\% \pm 1.3\%$ & no covering on forehead, specialized hardware \\ \hline
\end{tabular}%
}

\end{table*}

\subsection{Gait biometrics}
Gait pattern is one of the most popular biometric traits ~\cite{cunado1997using}, but sensing of gait pattern is a challenge. An early work on using pose agnostic gait recognition~\cite{goffredo2009self} used a view-invariant video sequence, which paved the way for many works in the future. The use of 3D information for gait biometrics was proposed~\cite{ariyanto2011model} in which kinematic models were learnt from human motion in voxel space, which was later used for classification. Though gait biometrics seems to be a difficult security measure to attack, it was proved that with clothing variation, gait patterns can be spoofed~\cite{hadid2012can}. This research began on making gait recognition invariant of clothing and luggage. Getting gait information from mobile sensors such as accelerometers and gyroscope~\cite{zhong2015pace} paved the way for non-visual approaches for gait biometrics. The trend of recent surveys in a chronological fashion~\cite{wan2018survey,isaac2019trait,makihara2020gait,sepas2022deep,parashar2023deep} shows that the field of Gait biometrics is gradually moving towards the use of deep learning models with a monocular camera using 3D vision as the only sensor. We have compiled some recent works in 3D monocular gait recognition in Table~\ref{tab_gait} that shows the trend of gait as a biometric using a monocular camera as the only sensor. Time series analysis of skeleton points~\cite{toral2022gait}, many times enhanced with depth~\cite{favorskaya2022accurate} and optical flow~\cite{khan2023hgrbol2} are mostly used.

\begin{table*}[]

\caption{Recent works in Gait biometrics using normal monocular camera}
\label{tab_gait}
\resizebox{\textwidth}{!}{%
\begin{tabular}{|p{15mm}|p{80mm}|p{40mm}|p{50mm}|p{25mm}|}
\hline
\textbf{Work} & \textbf{Technique} & \textbf{Dataset} & \textbf{Accuracy} & \textbf{Challenge} \\ \hline
Hua~\etal~\cite{hua2022gait} & Fusion of global long-term attention (GLTA) and local short-term attention (LSTA) for extracting personal static and dynamic physiological features, novel architecture called physiological feature extraction (PFE) network to concatenate gait silhouette and physiological features for classification & CASIA-B~\cite{yu2006framework}, Multi-state Gait dataset (MSGD) created like MVLP gait dataset~\cite{takemura2018multi} with $60$ persons with $14$ angles with $14$ sequences in each angle per person & rank 1 accuracy with Normal walking, carrying backpack, wearing coat on CASIA-B $96.5\%$, $92.6\%$, $79.8\%$ on MSGD $91\%$, $85.9\%$, $75.3\%$ & accuracy dips when carrying accessories\\ \hline
Favorskaya ~\etal~\cite{favorskaya2022accurate} & The depth is estimated from a U-net type architecture and then the foreground is extracted where optical flow is computed, then the interpolated motion is used for skeleton point time trajectory estimation, which are the features for classification & TUM-GAIT dataset~\cite{hofmann2014tum} & best accuracy, normal $99.7\%$, with bag $68.3\%$ and with different shoes $98.1\%$ & only suitable for particular camera pose\\ \hline
Khan~\etal~\cite{khan2023hgrbol2} & Optical-flow based region extraction for transfer learning on EfficientNet-B0, Bayesian optimization for enhancing video frames, feature fusion by named-Sq-Parallel Fusion then Entropy controlled Tiger optimization used for classifying using Extreme Learning Machines & CASIA-B~\cite{yu2006framework}, CASIA-C~\cite{arshad2022multilevel} & rank 1 accuracy on CASIA-B Normal walking, carrying backpack , wearing coat  $96.82\%$, $94.1\%$, $85.21\%$, CASIA-C normal walk with bag, slow walk, normal walk, fast walk $99.4\%$, $92.8\%$, $94.1\%$, $93.6\%$ & fails when high inter-class similarity is there\\ \hline
Toral~\etal~\cite{toral2022gait} & Skeleton extraction by openPose~\cite{cao2017realtime} and then using Linear Discriminant Analysis and Quadratic Discriminant Analysis on time series of skeleton points & Dataset created from $14$ individuals walking $4$ m  distance $10$ times each, camera framerate at $30$ FPS and a test train ration of $7:3$ & Accuracy of $0.9733$ and $F_1$ of $0.9732$ & camera has pose bias in dataset\\ \hline
\end{tabular}%

}

\end{table*}

\subsection{Eye movement biometric}
\label{lab_eye}
One of the earliest surveys works for measuring eye movement and its use in behaviour classification~\cite{young1975survey} proposed the way for using it as a biometric trait. The 3 dimensions of eye movement are the 2D spatial locus on a 1D temporal dimension. For many years the area remained dormant till eye movement, as a biometric, was established ~\cite{kasprowski2004eye}. The work argued that eye movement biometrics is a short-term recognition process with a maximum duration of $10$ seconds, but later works proved that with more time accuracy becomes better~\cite{jia2018biometric,abdelwahab2022deep} at the cost of high execution time. They also showed how deep learning approaches can be used for better performance. Eye movement biometrics has been a relatively less used biometric trait due to the requirement of specialized hardware (high-speed monocular camera) and long recording times. Early survey works like~\cite{goudelis2008emerging} prophesied in favour of eye movement as a rising trait in biometrics. Recently with advancements in deep learning models and sensor technology, the interest in the field has rekindled. A thorough survey~\cite{brasil2020eye} has shown how eye movement biometrics have gained pace and another work~\cite{schroder2020robustness} showed how the robustness of the method has increased through time using deep learning approaches. Some recent surveys~\cite{mahanama2022eye,startsev2022evaluating} have shown a trend of using normal monocular cameras in the research community, but it is still behind the accuracy of high-speed eye movement trackers. We have analysed some recent works on eye movement biometrics with a monocular eye tracker as the sensor in Table~\ref{tab:tab_eye movement} and have discovered that there is a dearth of datasets in the area. We also observed that accuracy drops with low-cost low-speed trackers thus keeping the problem of eye movement biometrics with a normal monocular camera still open. CNN is the most preferred methodology on the 2D coordinate locus~\cite{lohr2022eye,lohr2022eye2,raju2022iris,yin2022effective} on all recently explored work.

\begin{table*}[]
\caption{Recent works in eye movement as biometric using high-speed monocular eye trackers}
\resizebox{\textwidth}{!}{%
\label{tab:tab_eye movement}
\begin{tabular}{|p{17mm}|p{20mm}|p{65mm}|p{20mm}|p{40mm}|p{25mm}|}
\hline
\textbf{Work} & \textbf{Sensor} & \textbf{Technique} & \textbf{Dataset} & \textbf{Accuracy} & \textbf{Challenge} \\ \hline
Lohr~\etal~\cite{lohr2022eye} & EyeLink 1000 eye tracker at a 1000 Hz sampling rate & End-to-end Eye Movement Biometrics (EMB) with a novel variant of parameter efficient Densenet & Gazebase~\cite{griffith2021gazebase} and Judo1000~\cite{makowski2020biometric} & EER $0.32\% \pm 0.04\%$ best at $500Hz$ with $5 \times 12$ s duration, EER $2.67\%$ with same settings & Needs high-speed monocular eye-tracker\\ \hline
Lohr~\etal~\cite{lohr2022eye2} & EyeLink 1000 eye tracker at a 1000 Hz sampling rate & Exponentially-dilated CNN with $440$ times less learnable parameters, trained with multi-similarity loss, that works on low cognitive tasks & Gazebase~\cite{griffith2021gazebase} & HSS($1000Hz$) EER $0.1125$ with $1$ round, TEX($1000Hz$) EER $0.2110$ with $6$ rounds & degraded performance on low-cost eye trackers\\ \hline
Raju~\etal~\cite{raju2022iris} & CMITECH BMT-20 iris imager & ResNet 16 based approach to classify printed iris and genuine by eye movement with very low recording duration & ETDPADv2 ~\cite{rigas2015eye} & $1500$ ms timed SAS-I EER $3.62\%$, ACR $98.06\%$, APCER $0.14\%$, NPCER $3.75\%$, ACER $1.94\%$ & Does not consider replay and lens attacks\\ \hline
Yin~\etal~\cite{yin2022effective} & EyeLink 1000 eye tracker at a 1000 Hz sampling rate & Motion Information (MI) and Saccade Distribution Map (SDM) feature extraction network, each supplemented by 2 CNNs, concatenated feature train on L2 used for classification & Gazebase~\cite{griffith2021gazebase} & HSS-12C $5.25\%$ (only MI), RAN-12C $6.3\%$ (SDM+MI) TEX-12C $7.33\%$ (only MI) & Needs long exposure time of $12$s\\ \hline
\end{tabular}%
}
\end{table*}

\vspace{-2mm}

\subsection{Other available works}
\vspace{-2mm}
Other than using monocular cameras and standard traits, various other sensors and traits have emerged in recent times. 3D reconstruction of finger surface with multiple shots taken with a monocular camera~\cite{cui2023monocular, ramachandra2023finger} is a very recent trait in biometrics that has proved better accuracy than standard fingerprint and vein-pattern approaches. Another recent work has proved how this approach can provide better security against presentation attacks~\cite{purnapatra2023presentation}. In many cases, 3D information is obtained using comparatively expensive sensors. Structured light is been used many a time on the face for fast 3D information retrieval but that causes irritation to the eyes. Millimeter wave radar has been used to mitigate this issue~\cite{xu2022mask} and it proved a reliable way for face recognition and liveliness detection. The use of Lidar as the sensor for gait biometrics~\cite{alvarez2022biometric} has proved better accuracy due to its efficiency in detecting 3D features. Deep learning has been used on pen holding pose~\cite{wu2022towards} obtained from video clips, to add towards accuracy of live 2D signature biometrics. Monocular images of finger knuckles~\cite{singh2022biometric} have also been proved as a trait of biometrics with relatively low-cost cameras. Thermal cameras have been used to use heat signatures of hand for biometrics~\cite{knish2022thermal} and have high resilience against presentation attacks. The trajectory of VR controllers with respect to VR headset~\cite{miller2022combining} has been used as a biometric to incorporate the behaviour of persons using such hardware. Facial microexpressions are known to be a good trait in biometrics but a simple monocular camera often fails to capture the finer nuances required as features for it. In this direction, event cameras~\cite{becattini2022understanding} have been used recently to capture more details with a very high frame rate. Soft biometrics for person re-identification has been used widely but recently WiFi radio signature distortion due to human presence has been proved as a metric of soft biometrics~\cite{avola2022person}. Some recent datasets have come up, among whom, monocular mobile-phone camera face dataset with ears~\cite{freire2023zero}, monocular full body dataset from different angles and altitudes and different distances (as long as $1$ km) over a thousand subjects~\cite{cornett2023expanding}, soft biometric dataset from monocular cameras with gender, age and ethnicity~\cite{hassan2022onedetect} and dataset for 3D fingerprint presentation attack with monocular images~\cite{kolberg2023colfispoof} are some which deserves mention. Such trends clearly show that monocular camera used for various contact-less biometric traits is going to be a leading area of research in the near future. 

\subsection{Commercial solutions}
In terms of commercial solutions also many recent systems can be found, focusing their attention on biometrics with monocular cameras. Some of the key players are FaceTec Inc. with its selfie camera-based face recognition. It is a real-time 3D FaceScan that collects time-stamped, unreusable Liveness data and creates a 3D FaceMap \cite{2}. TBS 2D Iron is a robust fingerprint sensor that provides the user with a sense of comfort and reliability. It scans the 3D fingerprint from visual spectra camera. It is ideal for both indoor and outdoor applications and it can work efficiently in harsh environmental conditions to deliver superior identification performance \cite{3}. Fujitsu PalmSecure technology from Fujitsu Inc. is a palm vein-based authentication solution that utilizes industry-leading vascular pattern biometric
technology. The Fujitsu PalmSecure sensor uses near-infrared light to capture a person’s 3D palm vein pattern, generating a unique biometric template that is immediately encrypted in the sensor before transmission\cite{4}.

\subsection{Summary}
After analysis of recent trends in the field of biometrics using monocular cameras, the general patterns can be summarized into some basic points. The most important observation is that 3D face reconstruction from monocular images, using pseudo-depth map generation, is a growing trend, with the betterment of models through the years. Novel devices using normal camera on rotating mechanism for 3D reconstruction using SFM, on non-dynamic features is gaining steam. For using the monocular camera at its full potential, short videos with spatiotemporal attention are gaining popularity in anti-spoofing biometric approaches. It is also observed that the use of NIR cameras or normal cameras with NIR LEDs is gaining popularity for subcutaneous biometrics. Emerging technologies such as Lidar and event cameras are being used with their decreasing cost factors as an alternative to monocular cameras. Recently novel datasets are coming up in 3D face, finger and gait. The use of multimodal information like audio with video clips is getting attention (in both meanings of the word!).  

\section{Open research problems}
After going through the recent advancements in the field of biometrics using the monocular camera we have observed that there are a few open challenges remaining that needs to be addressed. Most of the challenges depend on further refinement of deep learning models and few are also related to enhancement of sensor hardware of a monocular camera. They are as follows:
\begin{itemize}
    \item \textbf{Lighting model invariant 3D reconstruction of traits from monocular images:} As observed in recent works, 3D trait reconstruction such as face from a single image depends a lot on the light distribution. This is a restriction in the adoption of a monocular camera as a universal sensor in any environment. Research is expected in this direction to accurately model 3D features of space irrespective of the lighting model.
    \item \textbf{Auto-preprocessing to nullify camera corrections: }Use of deep learning models on images and videos often suffers from unnecessary preprocessing imposed by commercially available, cost-efficient monocular cameras, as their primary purpose is to obtain visually aesthetic photographs. In the process of serving that, it alters the image quality, often to the extent that latent distinguishing features are attenuated beyond extraction. Models are required to be designed that can extract and enhance such features in an architecture and monocular camera-type agnostic manner.
    \item \textbf{Multipurpose filters with electronically adjustable pass frequency: }To use the same sensor in various traits, usually filters on the lens of the camera are enough. But those filers have a fixed cut-off frequency unlike their electrical counterparts working on electrical signals. It is expected that researchers will investigate the building of filters, whose cut-off frequency can be adjusted electrically, maybe by altering the alignment of crystals suspended in some medium.
    \item \textbf{Cost-effective faster frame rate with reasonable exposure: }Many traits of biometrics, especially eye movement and gait suffers from low frame rates available in standard cameras. Generally, the pixel size of sensors is enough and with proper software and sampling frequency much higher frame rates can be achieved. This is evident in smartphones where slow-mo camera apps can be installed without any change in sensor geometry. How to do such enhancements in a cost-effective manner, perhaps using edge computing, in real-time is an area where further research is expected for the discussed field.
    \item \textbf{Ethnicity and skin tone agnostic feature extraction:} This is perhaps one of the open challenges in biometrics that is still not properly solved. This is due to the bias in facial models and data points in training datasets\cite{das2018mitigating}. A more generalized approach to modelling biometric traits which is robust to any such coverlets is expected by researchers.
    \item \textbf{Generalized techniques: }Achieving generalized featuring and modeling will always in the to-do-list. In this interoperability of the models with respect to acquiring sensors is an important aspect to study. Another important aspect in this direction will be to generalise attack detection across various traits of biometrics using a monocular camera. The types of attacks are increasing every day. To tackle newer attacks without having enough data points to train the models a generalized anomaly detection approach is expected. Such an approach can only be realized if macro features learnt by deep models are contextually represented in a human-interpretable manner. Thus bringing explainability to the classification and verification processes.
    \item \textbf{Explainable models:} For this avenue of research working on the explainable model can lead to improvement of the modelling 3D information from 2D data. Thereby, developing the overall performance of the pipeline of 3D monocular biometrics. 
    
\end{itemize}

\section{Conclusions}
This article reviews recent literature on 3D biometrics employing monocular vision and finds that research in this area has gained a recent surge. This is due to the power of deep learning that produced a platform that can nurture latent information in traditional monocular sensors and produce comparable results that we can achieve while using sophisticated 3D sensors. In this article, we reviewed recent techniques, showing that face, vein, eye movement and gait, are prevalent traits that are considered for 3D biometrics employing monocular vision. We discussed their benefits and limitations, as well as challenges with respect to biometric acquisition processing, and commercial systems available. Finally, we elaborated on some of the open research problems in the field.

{\small
\bibliographystyle{ieee}
\bibliography{egbib}
}

\end{document}